\documentclass[runningheads]{llncs}

\usepackage{eccv}

\usepackage{eccvabbrv}
\usepackage{amsmath}
\usepackage{pifont}

\usepackage{graphicx}
\usepackage{booktabs}
\usepackage[ruled,vlined]{algorithm2e}
\usepackage{appendix}

\usepackage[symbol]{footmisc}

\usepackage[accsupp]{axessibility}  

\newcommand*\samethanks[1][\value{footnote}]{\footnotemark[#1]}

\usepackage[colorlinks]{hyperref}

\usepackage{orcidlink}
\usepackage{wrapfig}
\usepackage{algcompatible}
\usepackage{soul}

\newcommand{\sysname}{TCAN}
\begin{document}

\title{\sysname: Animating Human Images \\
with Temporally Consistent Pose Guidance \\
using Diffusion Models}



\titlerunning{TCAN: Temporally Consistent Human Image Animation}
\author{Jeongho Kim\inst{1}\thanks{Authors contributed equally to this work.}\orcidlink{0000-0003-4058-8163} \and
Min-Jung Kim\inst{1}\samethanks\orcidlink{0000-0003-3799-8225} \and
Junsoo Lee\inst{2}\orcidlink{0009-0006-3467-7855} \and 
Jaegul Choo \inst{1}\orcidlink{0000-0003-1071-4835}\\
}

\authorrunning{J.~Kim and M.~Kim et al.}


\institute{
    ~\inst{1}KAIST \ \ ~\inst{2}NAVER WEBTOON AI
    \\
    \email{\{rlawjdghek, emjay73, jchoo\}@kaist.ac.kr, ljs93kr@gmail.com}
}

\maketitle
\begin{abstract}
Pose-driven human-image animation diffusion models have shown remarkable capabilities in realistic human video synthesis.
Despite the promising results achieved by previous approaches, challenges persist in achieving temporally consistent animation and ensuring robustness with off-the-shelf pose detectors.
In this paper, we present~\sysname, a pose-driven human image animation method that is
robust to erroneous poses and consistent over time.
In contrast to previous methods, we utilize the pre-trained ControlNet without fine-tuning to leverage its extensive pre-acquired knowledge from numerous pose-image-caption pairs.
To keep the ControlNet frozen, 
we adapt LoRA to the UNet layers, 
enabling the network to align the latent space between the pose and appearance features. 
Additionally, by introducing an additional temporal layer to the ControlNet, 
we enhance robustness against outliers of the pose detector. 
Through the analysis of attention maps over the temporal axis, we also designed a novel temperature map leveraging pose information, allowing for a more static background.
Extensive experiments demonstrate that the proposed method can achieve promising results in video synthesis tasks encompassing various poses, like chibi.
Project Page:  \href{https://eccv2024tcan.github.io/}{https://eccv2024tcan.github.io/}

%

\end{abstract}
\section{Introduction}
\begin{figure}[t]
    \centering
    \includegraphics[width=\textwidth]{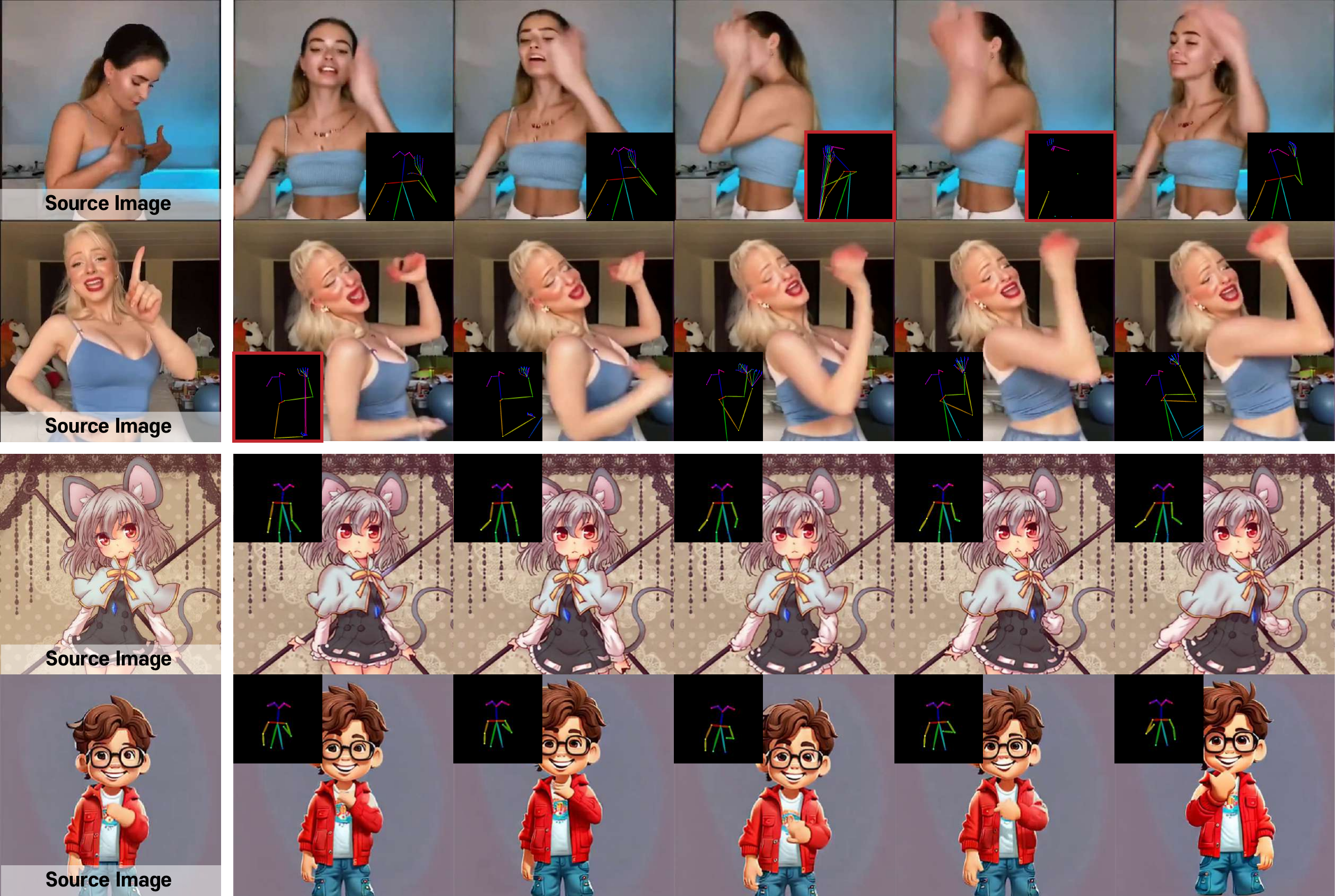}
    \captionof{figure}{
    Generated results of~\sysname: The first two rows show the results on the TikTok dataset, and the last two rows show the results on chibi animation characters.
    All results are generated using~\sysname~trained on the TikTok dataset. Severely erroneous input poses are highlighted in red. Note that the proposed~\sysname~can generalize to poses with outliers and unusual ratios, such as those of chibi characters.}
\end{figure}
Pose-driven human-image animation is a task that breathes life into a static human in an image.
Given a source image and a driving video, the task aims to transfer the motion in the driving video to a human in the source image. 
Although it is an intriguing task because of its wide impact on many applications such as social media, entertainment, and movie industry, several factors make this task challenging.
First, the identity in the source image should remain the same over time, changing only the pose.
In other words, the appearance of a human in the generated video should not be affected by the appearance of a human in the driving video.
Secondly, the occluded regions in the source image should be naturally inpainted as the pose changes, 
while ensuring that the inpainted regions remain unchanged over time. 

One possible approach is to impose a pose-guided image generation model frame-by-frame~\cite{zhang2023adding, zhao2024uni, ye2023ip}. 
Upon the success of the Text-to-Image (T2I) diffusion model, 
researchers made an effort to control the diffusion model using additional conditions 
in addition to the 
text prompts~\cite{ye2023ip,zhang2023adding,zhao2024uni,mou2023t2i,zhu2023tryondiffusion, cheong2023upgpt, cheong2023visconet}.
ControlNet~\cite{zhang2023adding} is the one representative work that enriches the controllability of the stable diffusion model, embracing various conditions, including pose, depth, and edge map.
Uni-ControlNet~\cite{zhao2024uni} extends ControlNet to handle multiple conditions at a time,
enabling the generation of an image conditioned on a source image and a pose image.
Meanwhile, in an effort to 
adapt the appearance of the source image along with text prompt,
IP-Adapter~\cite{ye2023ip} introduces an image prompt adapter that embeds the fine-grained image feature into a diffusion model.
As the IP-Adapter is compatible with pose ControlNet~\cite{zhang2023adding}, 
a human image with a given pose can be generated using both.
However, the aforementioned methods are oriented towards a single image generation, 
not considering temporal consistency.

To incorporate temporal consistency into the diffusion model,
T2V models focus on the temporal alignment of latents by inserting temporal layers into the diffusion model.
In general, T2V models can be extended to Image-to-Video (I2V) models by replacing text conditions with CLIP image embedding~\cite{blattmann2023stable} or by concatenating an encoded initial frame to the temporal layer input along the channel dimension~\cite{blattmann2023align}. 
However, 
controlling the object in the generated video using a pose sequence is out of their research scope.
To control humans in the video, there have been several attempts to generate video conditioned on a pose sequence~\cite{siarohin2019first, yang2023effective, xu2024magicanimate, wang2023disco, hu2024animate, chen2023control}. MagicAnimate~\cite{xu2024magicanimate}, which is the most similar work to ours, extracts pose features from fine-tuned ControlNet and feeds them into the skip connection of Stable Diffusion UNet.
Since MagicAnimate fine-tunes the ControlNet with the dataset of the target domain, 
it tends to overfit to the target domain poses.
Meanwhile, using the pre-trained ControlNet without fine-tuning does not guarantee generalizability or target domain performance, 
because there is no way to align the appearance and pose features when the ControlNet is frozen.

To leverage the pre-acquired knowledge of ControlNet, 
we keep ControlNet frozen and train additional LoRA layer 
to align the appearance and the pose feature without compromising the ControlNet's performance.
Moreover, to ensure the background remains static while maintaining the dynamic motion of the foreground region,
we introduce a pose-driven temperature map to obtain the static background.
Finally, to enhance temporal robustness against the noise of off-the-shelf pose estimators, 
we integrate temporal layers into ControlNet for pose conditioning.

To sum up, our contributions are as follows:
\begin{itemize}
    \item We propose~\sysname, a novel human image animation framework based on the diffusion model that maintains temporal consistency and generalizes well to unseen domains.
    \item 
    We are the first to freeze the pre-trained ControlNet and find it helpful for not only preventing the network from overfitting to the training domain but also disentangling appearance from the pose. 
    By introducing the Appearance-Pose Adaptation (APPA) layer, we effectively address artifacts caused by frozen ControlNet adaptation. 
    \item We propose Pose-driven Temperature Map (PTM) and Temporal ControlNet (T-CtrlN) to enhance stability at the video level, demonstrating superior performance improvement in the video quality metrics (FID-VID and FVD).
    
\end{itemize}


\section{Related work}
\subsection{Controllable Diffusion Model for Image Generation}
Thanks to the scalability and stable training of diffusion models~\cite{ho2020denoising,song2019generative}, generative foundation models~\cite{rombach2022high,ramesh2021zero,saharia2022photorealistic} are now leading recent image generation research.
Building on the foundation model, active research has been conducted on the image-based controllable diffusion model. 
In particular, several studies have expanded the T2I foundation models into an image-controllable generative model through modules using 
image conditioning~\cite{zhang2023adding,li2023gligen,cheong2023upgpt,cheong2023visconet
,
ye2023ip,mou2023t2i,yang2023paint}. Such modules, trained on large-scale datasets, work as plug-and-play components.
They can be combined with other trained models and used 
across various applications. 
Despite their 
outstanding adaptation abilities at the image level, 
the aforementioned methods focus only on the controllability of a single frame, rather than a video. In this paper, we leverage modules trained on a vast amount of image datasets and extend them to the video level, presenting a video generation model that demonstrates enhanced generalization performance.

\subsection{Pose-driven Video Diffusion Models}
Several efforts~\cite{guo2023animatediff, singer2022make, blattmann2023align, blattmann2023stable} have expanded image generation models to video generation models, with most employing a two-stage training scheme~\cite{guo2023animatediff, blattmann2023align}. 
In the first stage, the image generation model is fine-tuned using video frames.
In the second stage, temporal layers are added and trained while keeping the base image generation model frozen. 

Recent pose-driven video diffusion models follow the same two-stage training paradigm. 
In the first stage, the pose-driven image generation model is trained on individual video frames and corresponding pose images extracted using a pose estimator.
Then, the model is inflated by inserting temporal layers to capture the temporal information of the pose sequence extracted from a driving video. 
Disco~\cite{wang2023disco}, for example, 
takes three properties as input
— human subjects, backgrounds, and poses — to achieve not only arbitrary compositionality but also high fidelity.
MagicAnimate~\cite{xu2024magicanimate} achieves state-of-the-art performance with a parallel UNet architecture conditioned on the source image through an attention mechanism. 
While both works show impressive video quality in human image animation, 
they address neither overfitting nor the incompleteness of pose sequences. 

Unlike the aforementioned methods, 
we propose a more generalized model that performs well on datasets beyond the real human image domain by adapting prior knowledge from the pre-trained model.
Additionally, we use the temporal ControlNet in the second stage to address incomplete pose information.
To the best of our knowledge, both overfitting and pose incompleteness have not been explored in previous research.








\section{Method}
\begin{figure}[t!]
    \centering
    \includegraphics[width=1\linewidth]{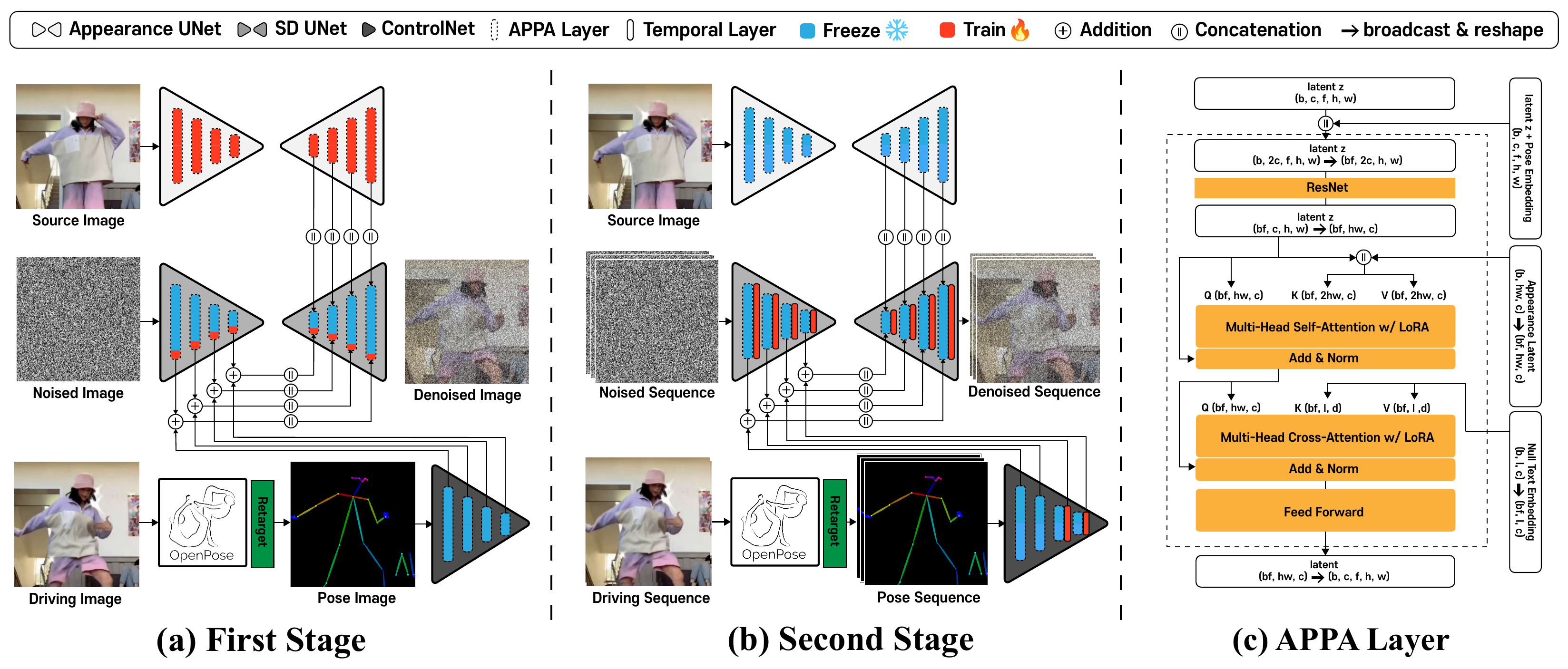}
    \caption{Our method involves a two-stage training strategy. We randomly select two images from the training video 
    and use them as the source image and the driving image, respectively. 
    In the first stage, 
    the appearance UNet and the APPA layer are trained conditioning on the source image while the ControlNet is frozen.
    In the second stage, we train the temporal layers in the denoising UNet and ControlNet.}
    \vspace{-0.1cm}
    \label{fig:overview}
\end{figure}

An overview of~\sysname~is presented in Fig.~\ref{fig:overview}. Given a source image and a driving video with $F$ frames, we aim to generate a video that: 1) incorporates the foreground appearance from the source image, 2) follows the pose of the driving video, and 3) maintains consistency in the background of the source image.

As shown in~\cref{fig:overview},~\sysname~follows a two-stage training framework widely adopted in current diffusion-based video generation works~\cite{guo2023animatediff,blattmann2023align}: 
In the first stage, the source image is transformed to match the pose of each frame in the driving video at the image level. 
To achieve this, we tackle the human image
\newpage
\begin{wrapfigure}{r}{0.47\textwidth}
    \centering    
    \includegraphics[width=0.45\textwidth]{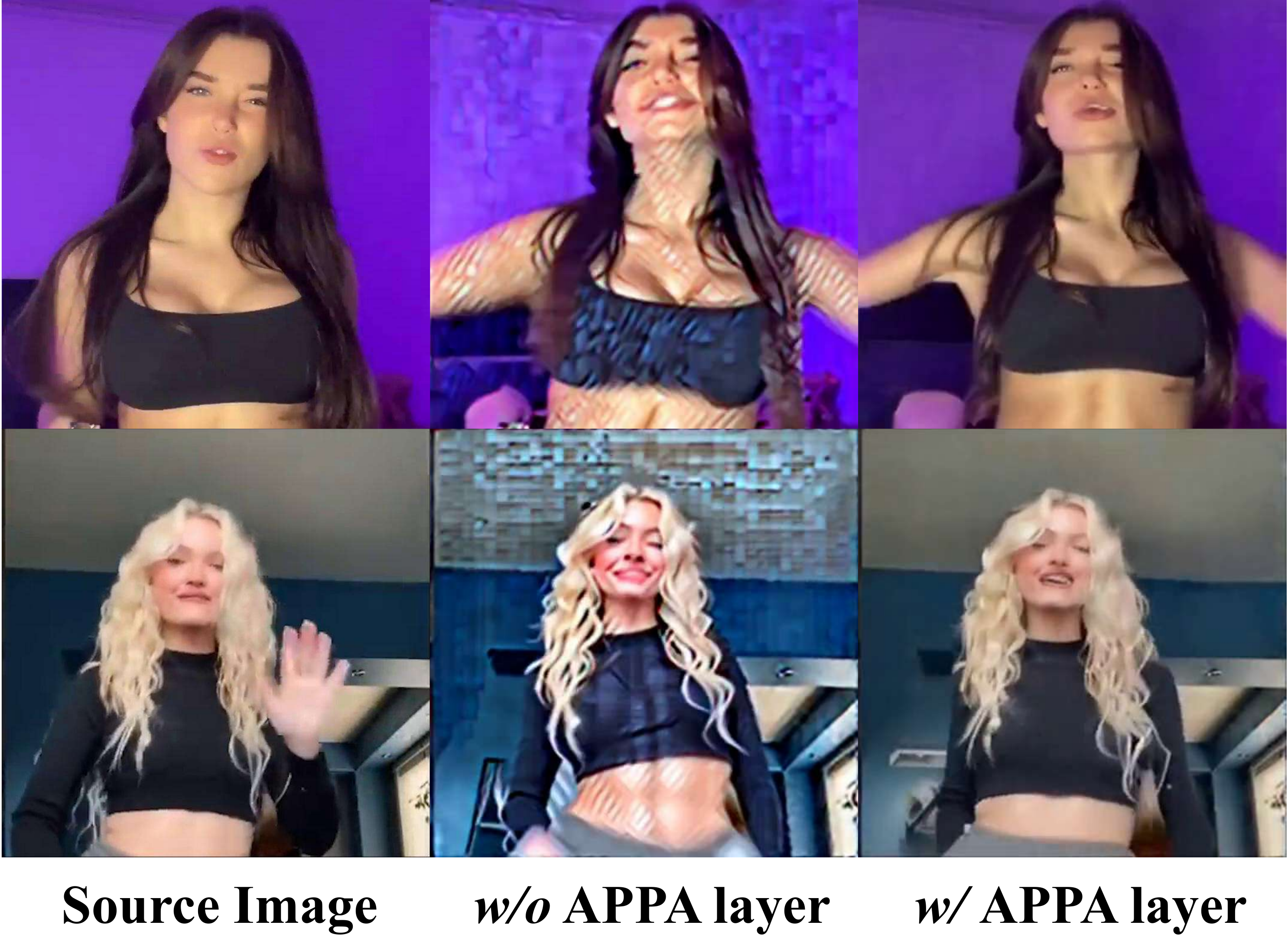}
    \caption{
    Ablation on the APPA layer. 
    Using only the frozen ControlNet significantly deteriorates texture quality, while the APPA layer accurately preserves the style of both the foreground and background from the source image in the output, showing notable differences.}
    \vspace{-0.5cm}
    \label{fig:APadaptation}
\end{wrapfigure}
\noindent animation task as a controllable 
image generation task under multiple image conditions (\textit{i.e.}, a pose condition and a source image), aiming to generate images that follow the style of the source image while adhering to the pose of the driving frames.  
We utilize OpenPose~\cite{cao2017realtime} to provide the pose information from the driving frames as input to the ControlNet. 
For the source image conditioning, we use an attention mechanism~\cite{xu2024magicanimate,zhu2023tryondiffusion} to pass the intermediate feature maps of the appearance UNet to the denoising UNet, providing finer features of the source image to denoising UNet. 
In the second stage, 
we extend the image generation model to a video generation model by adding temporal layers~\cite{guo2023animatediff}.

\subsection{Appearance-Pose Adaptation Layer}
To leverage the well-generalized pose conditioning capability of the pre-trained ControlNet, we opted to freeze the ControlNet in the first stage, in contrast to previous studies.
By adding the pose information from ControlNet to the denoising UNet's encoder feature $z$,
we embed the pose information 
into the denoising UNet.
This is similar to MagicAnimate~\cite{xu2024magicanimate}, but we use the feature from the frozen ControlNet, unlike MagicAnimate. 
Along with pose information, 
we inject the appearance information $z_a$ of the source image through the attention layer as in MagicAnimate~\cite{xu2024magicanimate}, which is formulated as the following equation:
\begin{equation}
    \begin{gathered} 
    \text{Attention}(\mathbf{Q},\mathbf{K},\mathbf{V}) = \text{softmax}\left(\frac{\mathbf{Q}\mathbf{K}^T}{\sqrt{d}}\right)\mathbf{V}, \\
    \mathbf{Q}=\mathbf{W}_0^Q z, \mathbf{K}=\mathbf{W}_0^K (z \parallel z_a), \mathbf{V}=\mathbf{W}_0^V (z \parallel z_a), 
    \end{gathered}    
\end{equation}
where $\parallel$ denotes the concatenation operation along the spatial dimension and $\mathbf{W}_0$ denotes the projection matrix of the denoising U-Net. 
We observe that solely training the appearance encoder with an adaptation of the pre-trained ControlNet 
allows for the successful reflection of the driving pose in the generated images. 
However, we also note a severe texture degradation in generation results, as shown in Fig.~\ref{fig:APadaptation}. 
We conjecture that the degradation of texture information is attributable to 
the misalignment between the pose information from the pre-trained ControlNet and newly trained appearance information.
Therefore, we propose an APpearance-Pose Adaptation layer (APPA layer), 
which preserves the appearance of the source image while maintaining pose information from the frozen ControlNet by aligning the feature of two different properties. 
Our proposed APPA layer is illustrated in~\cref{fig:overview} (c).
We implement the APPA layer by employing low-rank adaptation (LoRA) to the existing attention layers of the denoising UNet, specifically formulating the attention layer into the following equation, where appearance information $z_a$ is injected:
\begin{equation}
    \begin{gathered} 
        \mathbf{Q}=(\mathbf{W}_0^\mathbf{Q} + \Delta \mathbf{W}^Q) z, \mathbf{K}=(\mathbf{W}_0^K + \Delta \mathbf{W}^K) (z \parallel z_a), \mathbf{V}=(\mathbf{W}_0^V + \Delta \mathbf{W}^V) (z \parallel z_a). 
    \end{gathered}
\end{equation}
Here, $\Delta \mathbf{W}^{\{Q,K,V\}} = \mathbf{B}^{\{Q,K,V\}}\mathbf{A}^{\{Q,K,V\}}.$ The matrices $ \mathbf{B}\in\mathbb{R}^{d\times r}$ and $\mathbf{A}\in\mathbb{R}^{r\times c}$ denote low-rank-parameterized matrices with rank $r$. 
The appearance UNet and LoRA layers are trained in the first stage. 
By introducing the APPA layer, we effectively address artifacts caused by the frozen ControlNet adaptation. Moreover, we found that such an approach is helpful for preventing the network from overfitting to the training domain and disentangling appearance from the pose.

\subsection{Temporal ControlNet}
Pose-guided human image animation methods, including our~\sysname, are
conditioned on the pose sequence estimated from the driving video.
Consequently, the quality of the generated outputs relies on the accuracy of the pose estimator, 
making them inherently susceptible to inaccuracies in the predicted poses. 
One potential method to mitigate the impact of errors from the pose estimator is 
to fix the false detections from the estimated poses 
before use. 
 However, distinguishing 
 noisy keypoints 
 from 
 true positives 
 or adding missing keypoints 
 is challenging and tedious. 
 Moreover, this frame-by-frame noise removal approach fails to 
 ensure 
 temporal consistency across multiple frames of the pose sequence. 
 
 To prevent the generated video from collapsing due to abrupt and erroneous pose changes, 
 we propose Temporal ControlNet.
 Temporal ControlNet incorporates temporal layers, referred to as the motion module in AnimateDiff~\cite{guo2023animatediff}, which converts a T2I diffusion model into a T2V diffusion model.
 When the input tensor $z \in \mathbb{R}^{b\times c \times f\times h\times w}$ of $f$ frames is given, we reshape the tensor as $z \in \mathbb{R}^{bhw\times f\times c}$ and feed it to the temporal layer.
The output tensor from the temporal layer is reshaped as $z \in \mathbb{R}^{b\times c \times f\times h\times w}$, so that the next layer can extract features with respect to the spatial dimension.

\begin{figure}[t!]
    \centering
    \includegraphics[width=1\linewidth]{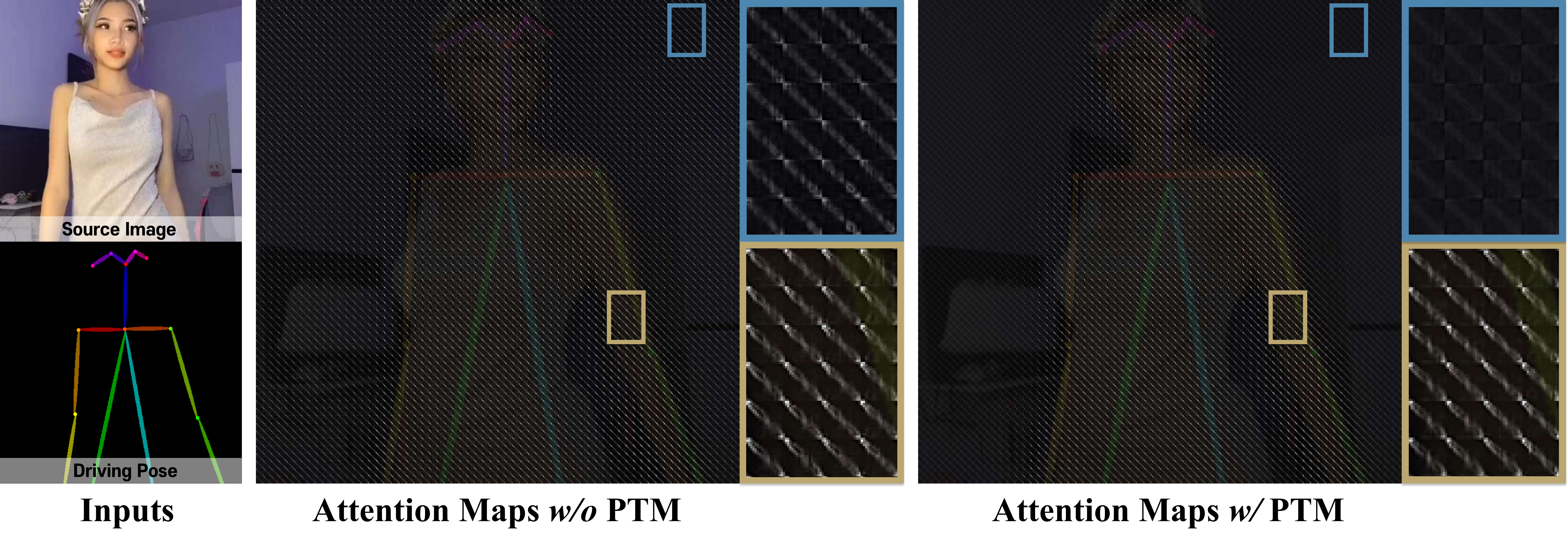}
    \caption{Visualization of the attention maps from the temporal transformer block at a resolution of $64\times64$. Best viewed when zoomed in.}
    \label{fig:PTM}
\end{figure}

The rationale behind this strategy is that when the ControlNet, pre-trained at the image level, receives a sequence containing erroneous poses as input, we want ControlNet to refrain from directly conditioning the denoising UNet on the erroneous poses. Instead, by introducing temporal layers into ControlNet, we can mitigate the erroneous poses by referencing neighboring correct poses along the temporal axis.
During training,
only the temporal layers are trained in the second stage, while the pre-trained ControlNet remains frozen.

\subsection{Pose-driven Temperature Map}
\label{sec:ptm}
Even with APPA layers and temporal ControlNet, 
the~\sysname~trained up to the second stage 
experiences flickering in the static region (i.e., background).
To address this issue, we propose a novel Pose-driven Temperature Map (PTM). 
The PTM is motivated by the observation that, \textit{compared to the foreground, the background tends to 
have a less specific focus 
over the temporal axis.} 
~\cref{fig:PTM}~supports the idea by visualizing 
the attention maps 
$m\in\mathbb{R}^{hf \times wf}$
of the temporal attention layer 
after the second stage training
, where $h$ and $w$ are 64 and the resolution of each attention map is $f \times f$. 
Each attention map represents the attention of each 
location 
along the temporal axis across consecutive video frames. 
For ease of interpretation, 
we resized the generated frames and the driving pose frame to a size of attention maps $m$ and overlaid them.
Interestingly, we observe that the attention map 
corresponding to the
moving parts, i.e., foreground objects, 
shows higher 
diagonal values compared to the static background. 
This indicates that dynamic objects stay at the same location \textit{temporally}, 
thereby referencing 
adjacent frames more than the static background region. 
Consequently, 
we designed a temperature map 
to smooth the attention scores far from the dynamic object. 

Given driving OpenPose sequence $\mathbb{P} = \{P^i \mid P^i \in \mathbb{R}^{3 \times H \times W}, \, i = 1, \ldots, f\}$ of $f$ frames, we first generate a binary mask $B\in\{0, 1\}^{H\times W}$ indicating the presence of OpenPose in each pixel.
Formally, this can be expressed as follows:
\begin{equation}
\begin{array}{c@{\hspace{0.5cm}}c}
    B(u,v) = \displaystyle\max_{i=1}^{f} B^i(u,v), 
    \ \ \ 
    B^i(u,v) = 
    \begin{cases}
        1 & \text{if } P^i(u,v) \neq (0,0,0) \\
        0 & \text{otherwise},
    \end{cases} 
\end{array}
\end{equation}
where $u$ and $v$ indicate the pixel coordinates. 
Let 
$B^\prime = \{ (u,v) | B(u,v)=1 \}$ 
be the set of pixel coordinates in $B$ with
a value of 1. Then, we can calculate the distance map $D \in \mathbb{R}^{H\times W}$ as follows:
\begin{equation}
    D(x,y) = 
    \frac{        
            \min \{\sqrt{ (x - u)^2 + (y - v)^2 } | (u,v) \in B' \}   
    }{\sqrt{(H/2)^2 + (W/2)^2}},
\end{equation}
where $x$ and $y$ indicate the pixel coordinates of distance map $D$. 
Based on the distance map, 
we smooth the attention map corresponding to regions 
far from 
the foreground objects
, i.e., the background. We define the PTM $\mathcal{T} \in \mathbb{R}^{H\times W}$ as follows:
\begin{equation}
    \mathcal{T} = \tau_{PTM} \cdot D + 1, 
\end{equation}
where $\tau_{PTM}$ serves as a hyper-parameter that adjusts the scale of the temperature values.
The value of 1 is added 
in order to maintain the attention map of the region near the OpenPose unchanged. 
Experimentally, $\tau_{PTM}=3$ shows reasonable performance, and we use this value throughout the paper. 
Finally, the PTM $\mathcal{T}$ is resized to match each temporal layer's spatial size in the denoising UNet and used as the temperature value for its attention operation. The formulation applied to the temporal layer of PTM $\mathcal{T}$ during inference stage is as follows (using \texttt{einops}~\cite{rogozhnikov2021einops} notation):

\begin{equation}
\begin{aligned}
    \mathbf{A^\prime} &= \texttt{rearrange}(\mathbf{Q}_t\mathbf{K}_t^T,\ \ \ \ \ \ \ \texttt{(b h w) f f} \rightarrow \texttt{b f f h w}) \notag \\
    \mathcal{T}^\prime &= \texttt{rearrange}(\mathcal{T},\ \ \ \ \ \ \ \ \ \ \ \ \ \ \texttt{b h w} \rightarrow \texttt{b 1 1 h w}) \notag \\
    \mathbf{A} &= \texttt{rearrange}(\mathbf{A^\prime}/(\mathcal{T}^\prime\sqrt{d}),\ \ \texttt{b f f h w} \rightarrow \texttt{(b h w) f f}) \notag \\
    z^\prime &= \text{softmax}(\mathbf{A})\mathbf{V}
\end{aligned}
\end{equation}
where $\mathbf{Q}_t=\mathbf{W}^Q_t z, \mathbf{K}_t=\mathbf{W}_t^K z, \mathbf{V}_t=\mathbf{W}_t^V z$. Here, $\mathbf{W}^Q_t, \mathbf{W}^K_t$, and $\mathbf{W}^V_t$ are weight matrices of the temporal layer. 



\renewcommand{\thefootnote}{\arabic{footnote}}
\subsection{Training and Long-Term Video Prediction}
\noindent\textbf{Training} 
We initialize the denoising UNet 
with 
the pretrained weights of Stable Diffusion 1.5\footnote{\label{SD15}Hugging Face SD 1.5: \href{https://huggingface.co/runwayml/stable-diffusion-v1-5}{https://huggingface.co/runwayml/stable-diffusion-v1-5}} 
, the appearance UNet with those of 
Realistic Vision\footnote{\label{RealisticVision}Civitai RealisticVision: \href{https://civitai.com/models/4201?modelVersionId=130072}{https://civitai.com/models/4201?modelVersionId=130072}}, and 
employ the weights of OpenPose ControlNet
\enlargethispage{3\baselineskip}
\footnote{\label{OpenPose}Hugging Face OpenPose: \href{https://huggingface.co/lllyasviel/sd-controlnet-openpose}{https://huggingface.co/lllyasviel/sd-controlnet-openpose}}.
In the first stage, we freeze all modules except the appearance UNet and the LoRA layer within the APPA layer. 
In the second stage, we introduce temporal layers to the denoising UNet and ControlNet while keeping the rest of the weights frozen.
The training objective function in both stages follows the simplified loss of the latent diffusion model~\cite{rombach2022high}.

\noindent\textbf{Long-Term Video Prediction}
Due to the memory constraints,~\sysname~cannot generate long-term video all at once. 
However, generating the video of $F$ frames by concatenating short videos of $f$ frames does not take temporal consistency into account. 
To address this, during inference, we apply MultiDiffusion~\cite{bar2023multidiffusion} along the temporal axis. 
This involves overlapping certain pose sequences in the model's input and averaging predicted noise for overlapping frames.
\section{Experiments}
\textbf{Evaluation setting}
During training, we use the TikTok dataset~\cite{jafarian2021learning} only.
For evaluation, we utilize two different datasets. 
First, we use 10 TikTok-style videos collected from DisCo~\cite{wang2023disco} as our test dataset, referring to it as the TikTok dataset evaluation. 
Second, to assess the 
generalization 
of~\sysname, we evaluate the model trained on the TikTok dataset using animation characters. 
Note that the textures and body proportions of the animation characters are quite different from those of real human datasets.
We select 10 images from the Bizarre Pose Dataset~\cite{chen2022transfer} as source images and use the 10 TikTok-style test videos as driving videos. We refer to this as the Bizarre dataset evaluation. 

\begin{figure}[t!]
    \centering
    \includegraphics[width=1\linewidth]{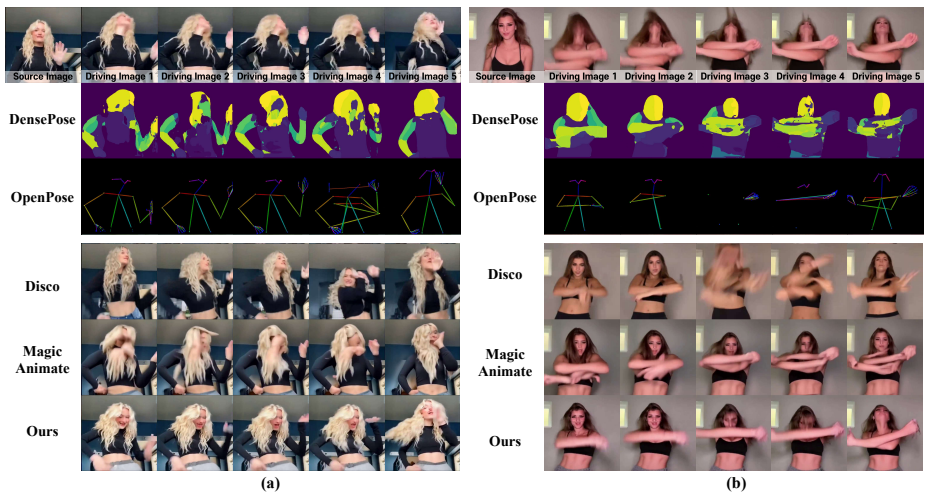}
    \vspace{-0.3cm}
    \caption{Qualitative comparison with baselines on the TikTok~\cite{jafarian2021learning} dataset. Frames from segments of the driving video are arranged sequentially.}
    \label{fig:qual_tiktok}
    \vspace{-0.5cm}
\end{figure}

\subsection{TikTok Dataset Evaluation} \label{sec:tiktok}
\noindent\textbf{Qualitative Results}
As shown in~\cref{fig:qual_tiktok}, we qualitatively compare our method with two baselines, DisCo~\cite{wang2023disco} and MagicAnimate~\cite{xu2024magicanimate}, both of which demonstrate state-of-the-art performance in human image animation. 
Note that MagicAnimate utilizes DensePose, while the others use OpenPose. 
Due to errors in the off-the-shelf pose extractor, 
significant pose information loss is evident in the fourth frame of ~\cref{fig:qual_tiktok}(a) and 
the third and fourth frames 
of ~\cref{fig:qual_tiktok}(b). 
As a result, DisCo generates an anomalous frame with severe blurring and artifacts (third frame of ~\cref{fig:qual_tiktok}(b)). 
Similarly, MagicAnimate, which is heavily dependent on DensePose, 
generates results that align with erroneous DensePose outputs (\cref{fig:qual_tiktok}(a)).
On the other hand, despite using the same OpenPose outputs as DisCo, ~\sysname~generates temporally consistent outputs without significant distortion or artifacts, especially in the fourth frame of~\cref{fig:qual_tiktok}(a). 

\noindent\textbf{Quantitative Results}
\begin{table}[t!]
\centering
    \resizebox{0.9\linewidth}{!}{
        \begin{tabular}{l|cccc|cc} 
        \toprule
        \multicolumn{1}{c}{Method} & L1$\downarrow$ & SSIM$\uparrow$  & LPIPS$\downarrow$ & FID$\downarrow$ & FID-VID$\downarrow$ & FVD$\downarrow$ \\ 
        \hline
        IPA \cite{ye2023ip}+CtrlN\cite{zhang2023adding} & 
            7.38E-04 & 0.459 & 0.481 & 69.83 & 
            113.31 & 802.44\\
        IPA \cite{ye2023ip}+CtrlN\cite{zhang2023adding}-V & 6.99E-04 & 0.479 & 0.461 & 66.81 & 86.33 & 666.27 \\
        DisCo \cite{wang2023disco} & 
            3.78E-04    & 0.668     & \underline{0.292} &  30.75 & 
            59.90       & 292.80    \\
        MagicAnimate & 
            3.13E-04 & 0.714 & \textbf{0.239} & 32.09 & 
            \underline{21.75} & \underline{179.07} \\
        \hline
        \textbf{\sysname} (\textit{w/ mm1}) & 
            \underline{8.85E-05} & \underline{0.734} & 0.299 & \textbf{29.07} & 
            29.74 & 189.77  \\
        \textbf{\sysname} (\textit{w/ mm2}) & 
            \textbf{8.81E-05} & \textbf{0.745} & 0.294 & \underline{29.35} & 
            \textbf{19.42} & \textbf{154.84}  \\
        \bottomrule
        \end{tabular}
    }
    \caption{Quantitative results on TikTok dataset evaluation. Metrics for baseline methods are cited directly from MagicAnimate~\cite{xu2024magicanimate}. The best performance is highlighted in bold, and the second best is underlined. \textit{mm1} and \textit{mm2} refer to version 1 and version 2 of the pretrained motion module from AnimateDiff~\cite{guo2023animatediff}. }
\label{tab:quan_tiktok}
\vspace{-0.5cm}
\end{table}
In~\cref{tab:quan_tiktok}, we compare our~\sysname~with four diffusion-based generation models.
The IPA + CtrlN is the model which combines IP-adapter~\cite{ye2023ip} and pre-trained OpenPose ControlNet~\cite{zhang2023adding},
and IPA + CtrlN-V is the model that extends IPA + CtrlN as a video generation model by inserting the temporal modules.
DisCo~\cite{wang2023disco} and MagicAnimate~\cite{xu2024magicanimate}
are recently proposed human image animation baselines.

We conduct TikTok dataset evaluation of 10 test videos 
using both image-level metrics (\textit{i.e.}, L1, SSIM~\cite{wang2004image}, LPIPS~\cite{zhang2018unreasonable}, and FID~\cite{heusel2017gans}) and video-level metrics (\textit{i.e.}, FID-VID~\cite{balaji2019conditional} and FVD~\cite{unterthiner2018towards}). 
Given test videos, the first frame of each video is used as the source image, and the remaining frames are used as the driving sequence. 
We use all frames of test videos without sampling during evaluation. 
For image quality assessment, 
we measure L1, SSIM, LPIPS, and FID between the generated and real frames. 
For video quality assessment metrics, FID-VID and FVD, we use 3D ResNet-50 and Inflated 3D ConvNet~\cite{carreira2017quo} pre-trained on the Kinetics dataset~\cite{kay2017kinetics} as feature extractors, respectively. 
For implementation, we follow the official DisCo repository
\footnote{\label{DisCo}DisCo Repository: \href{https://github.com/Wangt-CN/DisCo}{https://github.com/Wangt-CN/DisCo}}.

As reported in Table~\ref{tab:quan_tiktok},~\sysname~shows state-of-the-art performance in all metrics except LPIPS, 
without using the additional large-scale human dataset as in DisCo~\cite{wang2023disco} and MagicAnimate~\cite{xu2024magicanimate}.
We evaluate the performance using two versions of the motion module in AnimateDiff~\cite{guo2023animatediff} and observe that utilizing the advanced motion module (\textit{i.e.,} version 2) as the initial weight results in noticeably higher performance.


\begin{figure}[t!]
    \centering
    \includegraphics[width=1\linewidth]{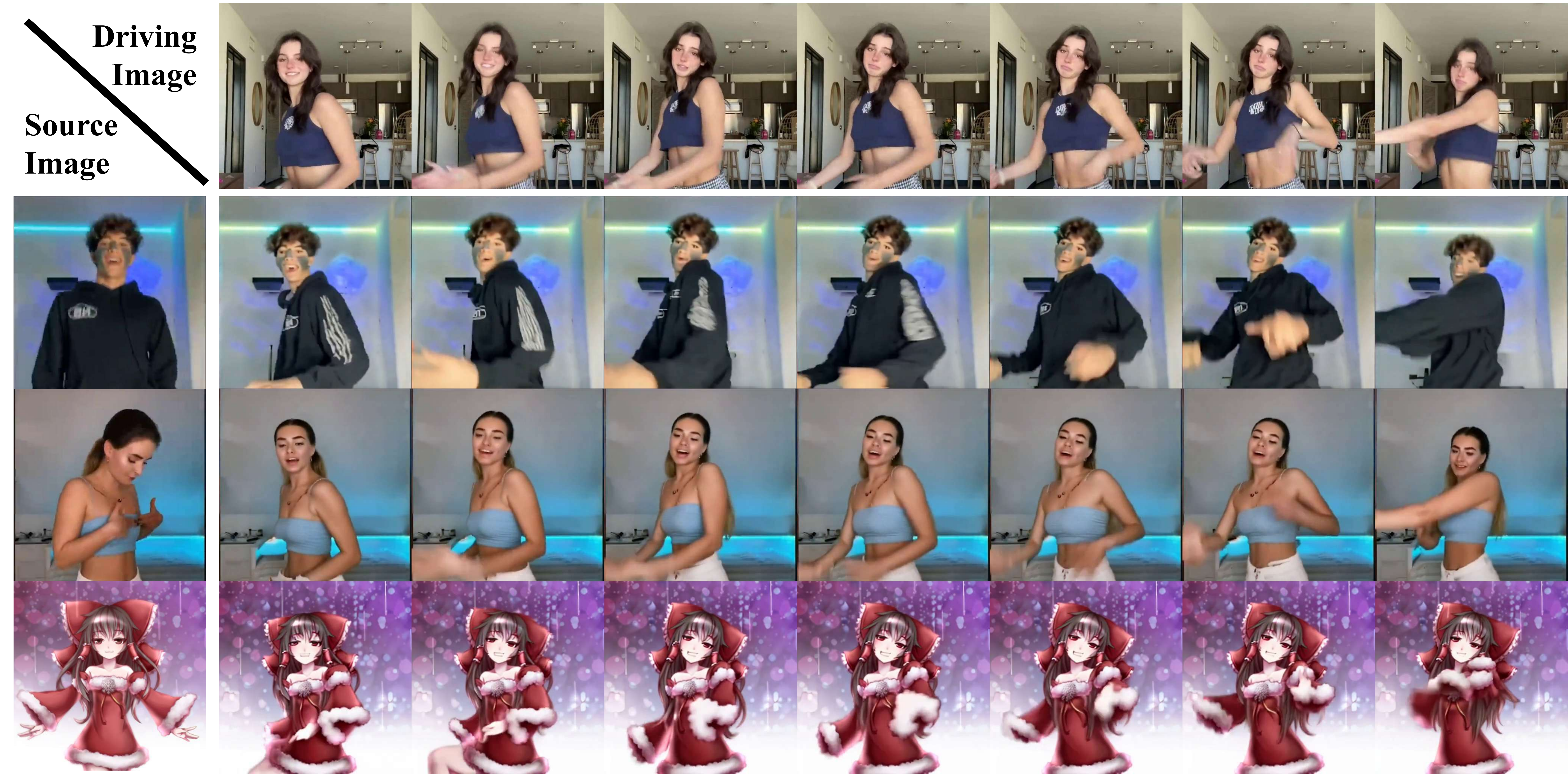}
    \caption{Qualitative results of~\sysname. This figure shows that our method effectively transfers motion information to various identities, including animated characters, despite differences in proportions between animation characters and humans.}
    \label{fig:qual_retarget}
\end{figure}

\subsection{Bizarre Dataset Evaluation with Pose Re-targeting}\label{sec:bizarre}
\begin{figure}[t!]
    \centering
    \includegraphics[width=1\linewidth]{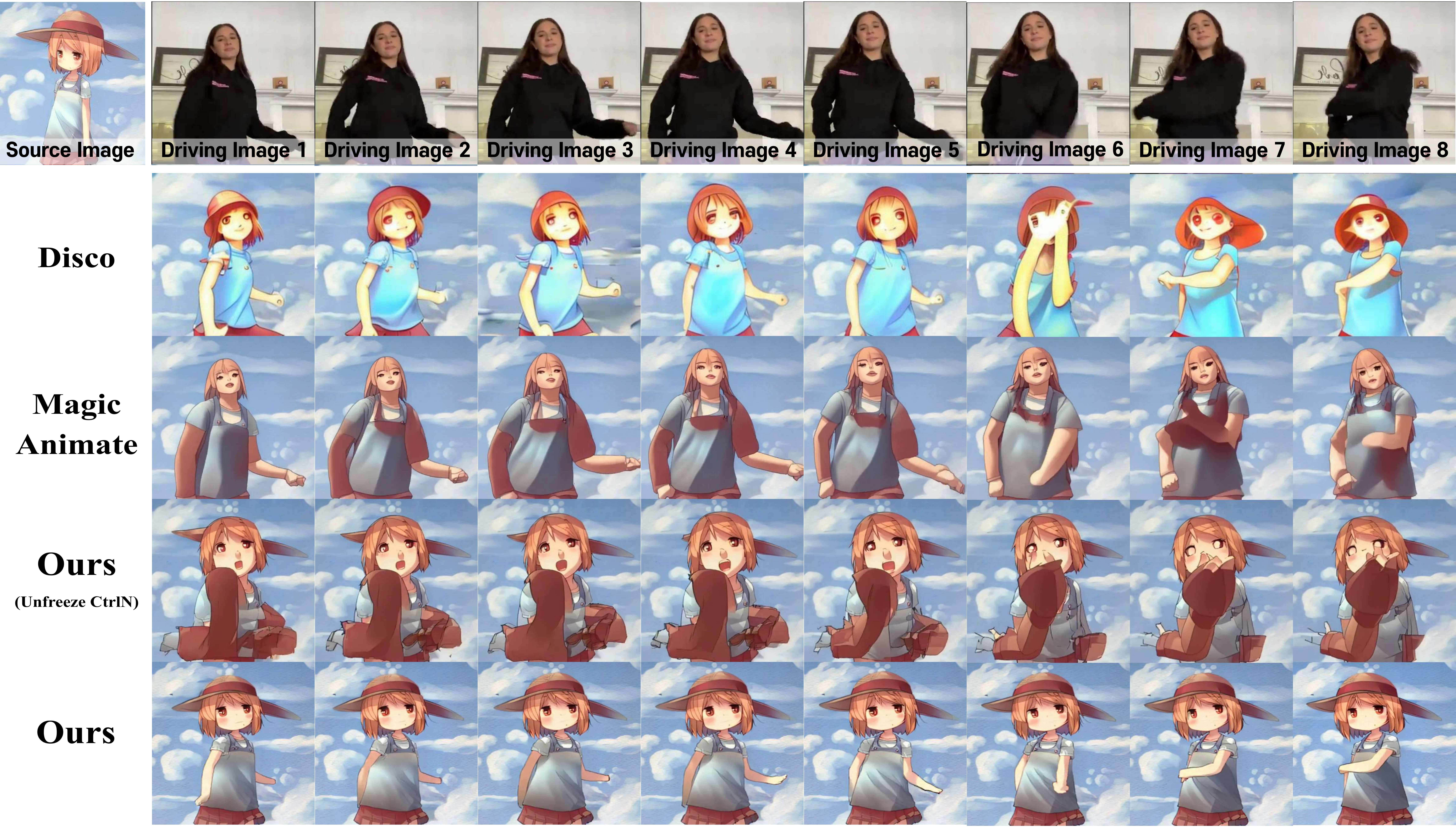}
    \caption{Qualitative comparison in a Bizarre dataset evaluation. As a source image, we use an animation character with different textures and body proportions compared to driving videos in TikTok dataset. Disco and our model, based on OpenPose are evaluated by applying the proposed pose re-targeting. To demonstrate the generalization performance of proposed method, we also visualize the results obtained when training ControlNet.}
    \vspace{-3mm}
    \label{fig:qual_bizarre}
\end{figure}

\noindent\textbf{Pose Re-targeting}
Beyond the TikTok dataset evaluation, where both the source image and driving sequence are derived from the same video, we extend our model to applications where the foreground object in the source image possesses body proportions different from those of an actual human. 
Specifically, when the source images are sampled from an animation domain,
there are
significant differences in body proportions compared to TikTok dataset, as depicted in~\cref{fig:qual_retarget} and ~\cref{fig:qual_bizarre}. 
To address this issue, we 
propose a re-targeting methodology. 
The pose re-targeting is designed to transfer the body the proportions of an object in the source image to that of the target frame, regardless of the object proportion in the driving sequence. 
When used in practice, we re-target the pose sequence estimated from the driving video so that the retargeted poses fit the proportion of the animation character in the source image.
As shown in Fig.~\ref{fig:qual_retarget}, re-targeting enables the preservation of proportions of an object in the source image, 
regardless of whether the object features a human or an animated character, 
while 
faithfully following the motions in the driving video.
Detailed explanations of the Pose Re-targeting are provided in the supplementary material. 

\noindent\textbf{Qualitative Results}
In~\cref{fig:qual_bizarre}, 
we qualitatively compare our~\sysname~with existing human image animation baselines, DisCo and MagicAnimate. 
For the sake of fairness, the re-targeted OpenPose sequence is used as a driving pose sequence for DisCo. 
We observe that the DisCo 
fails to maintain temporal consistency and character identity. 
Similarly, MagicAnimate, conditioned on 
DensePose~\cite{guler2018densepose}, generates videos that compromise the source image's identity with the shape in driving video. 
In contrast, thanks to the APPA layer and pose re-targeting, our model animates following the driving video's motion while preserving the source image's identity.
As demonstrated in the last two rows in~\cref{fig:qual_bizarre}, freezing the ControlNet plays a significant role in maintaining the identity. 
We conjecture that this occurs because freezing the ControlNet allows for the disentanglement of the network branches responsible for pose and appearance during training. Additionally, the pre-acquired knowledge of the ControlNet trained with OpenPose excels at understanding uncommon body postures and proportions. 
\\
\begin{wrapfigure}{r}{0.46\textwidth}
    \vspace{-1.2cm}
    \begin{center}
    \includegraphics[width=0.55\textwidth]{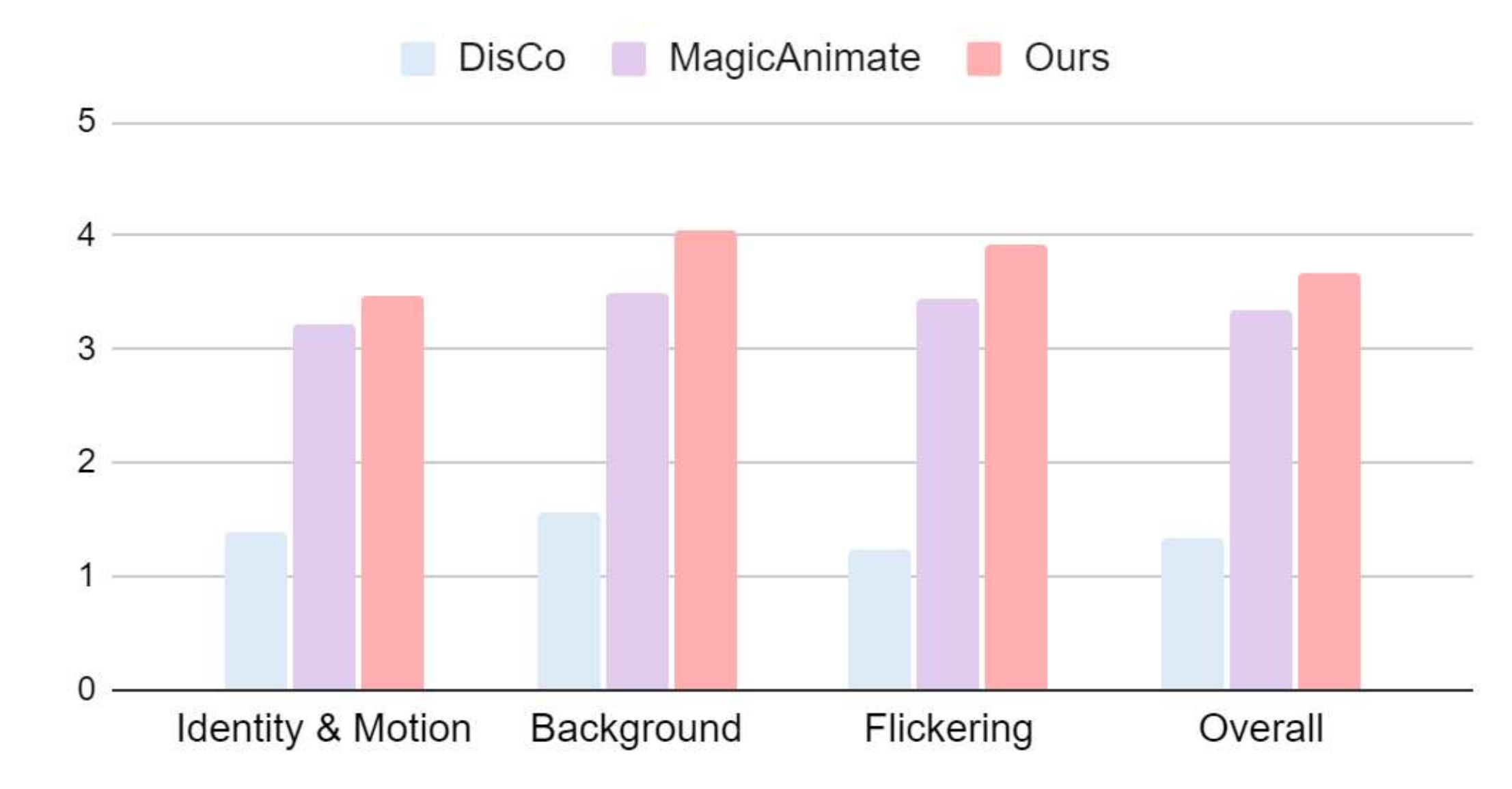}
    \vspace{-6mm}
    \caption{User study results involving a total of 27 participants.}
    \label{fig:user_study}
    \end{center}  
    \vspace{-1.7cm}    
\end{wrapfigure}
\noindent\textbf{User Study}
We further conduct a user 
study to evaluate the Bizarre dataset. 
A total of 27 participants were 
shown 10 videos generated by DisCo, MagicAnimate, and~\sysname, respectively. Following this, participants were asked to rate the video on a scale from 1 to 5 based on four criteria: 1) Motion and identity, 2) Background, 3) Flickering, and 4) Overall preference. The detailed instructions for each criterion are as follows:
\begin{itemize}
    \item Motion and Identity: \textit{Please rate how well the motion of the driving video was reflected while maintaining the identity of the source image.} 
    \item Background: \textit{Please rate the extent to which the background of the source image is maintained in the generated video.}
    \item Flickering: \textit{Please rate how minimal the flickering is in the video.}
    \item Overall: \textit{Please rate the overall preference score for the video.}
\end{itemize}

As shown in~\cref{fig:user_study}, our~\sysname~consistently outperforms previous state-of-the-art baselines. 
DisCo receives the lowest scores on all of the criteria due to severe flickering, identity loss, and artifacts in the foreground character. 
MagicAnimate also suffers from identity loss by generating characters that fit the DensePose of the driving video. 
Thanks to our novel PTM, our model demonstrates superior performance with minimal background flickering 
while maintaining the best performance in other criteria. Consequently, our model achieves the highest overall quality preference. 

\begin{figure}[t!]
    \centering
    \includegraphics[width=1\linewidth]{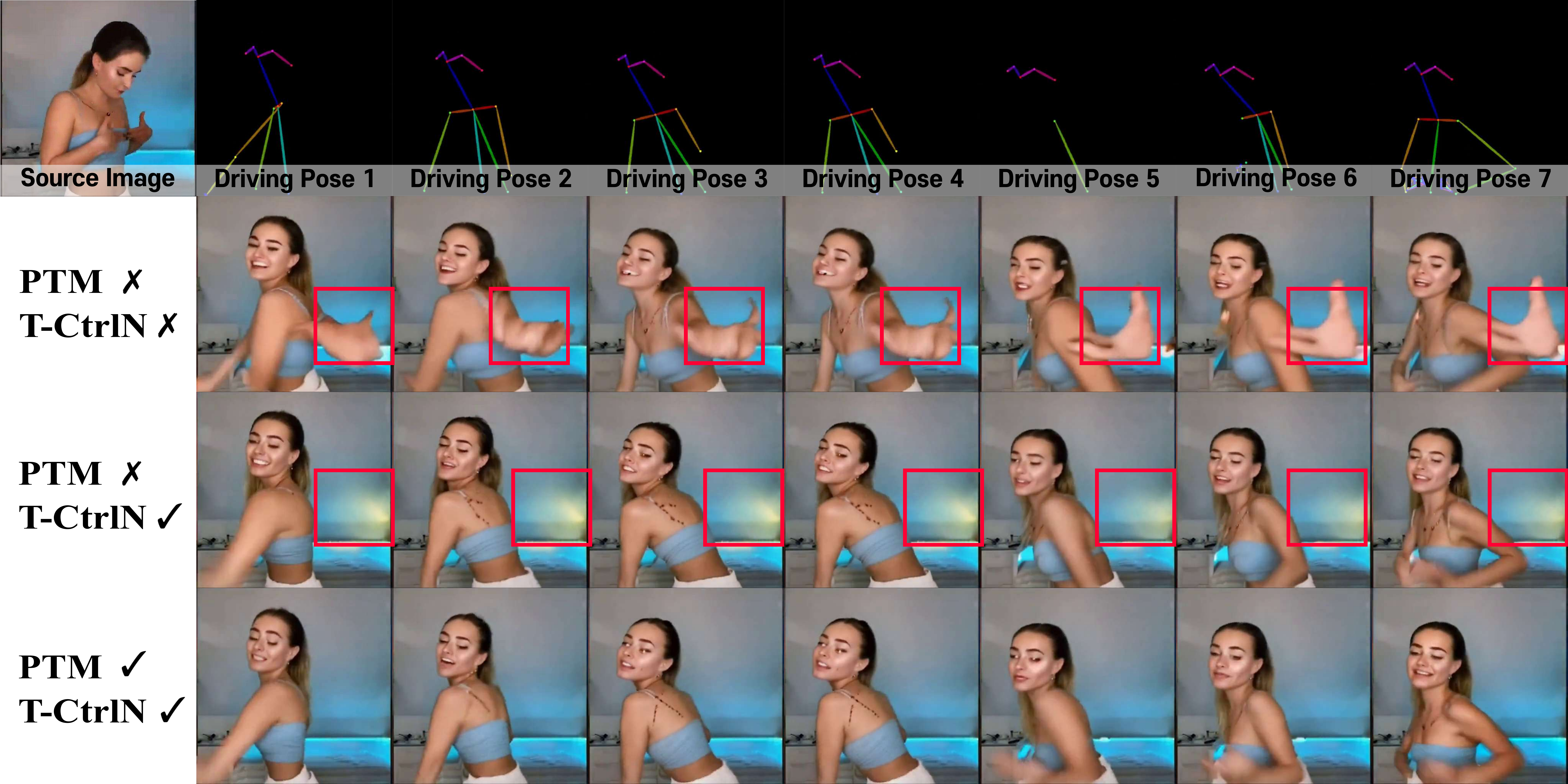}
    \caption{Qualitative results from our ablation study. We visualized the results of incorporating the proposed temporal ControlNet and PTM into the model with APPA layer. Red boxes highlight areas with body artifacts and background flickering.}
    \label{fig:qual_ablation}
\end{figure}
\section{Ablation Study}
\noindent\textbf{Qualitative Results}
We first investigate the generalization performance of APPA layer in the animation domain. We compare~\sysname~with a variant where ControlNet is unfrozen, and the APPA layer is omitted, For both models, we use re-targeted OpenPose as the input. As demonstrated in 
the~\cref{fig:qual_bizarre}, 
even when the ControlNet is initialized with pre-trained weights, 
the model exhibits a loss of prior pose knowledge 
if we unfreeze the ControlNet during training. 
Therefore, it fails to incorporate the variant poses in driving videos, leading to artifacts, particularly in the arm and torso regions. In contrast, our model, which keeps ControlNet frozen with the APPA layer, effectively maintains the appearance of the source image while accurately following the pose of the driving sequence.

Also, we examine the effect of our 
temporal ControlNet, and PTM.~\cref{fig:qual_ablation} shows eight consecutive frames from the generated video 
for each experimental scenario.  
First, we demonstrate the effect of the temporal ControlNet, which is shown in the second and third rows of~\cref{fig:qual_ablation}. 
The ControlNet branch expanded with temporal layers 
smooth out erroneous poses 
compared to the 
original ControlNet, 
where the right(orange) and left(green) arms are switched after the second frame 
and get back to normal at the last frame. 
This improvement is attributed to 
the temporal ControlNet, which effectively addresses outlier poses compared to the time-independent pose conditions provided by the image-level ControlNet model. Furthermore, as observed in the last row of Fig.~\ref{fig:qual_ablation}, the complete model with PTM effectively removes flickering artifacts caused by the yellow light, which is not present in the source image.
\begin{table}[t!]
\centering
    \resizebox{0.9\linewidth}{!}{
    \begin{tabular}{ccc|ccccccc} 
    \toprule
    \begin{tabular}[c]{@{}c@{}}
    \textbf{APPA}\\\textbf{Layer}\end{tabular} & \begin{tabular}[c]{@{}c@{}}\textbf{Temporal}\\\textbf{ControlNet}\end{tabular} & \textbf{PTM} & L1$\downarrow$       & SSIM$\uparrow$  & LPIPS$\downarrow$ & FID$\downarrow$   & FID-FVD$\downarrow$ & FVD$\downarrow$
    \\ 
    \hline
    \ding{55} & \ding{55} & \ding{55} & 1.13E-04 & 0.709 & 0.342 & 40.60 & 37.33 & 368.71 \\
    \ding{51} & \ding{55} & \ding{55} & 9.52E-05 & 0.728 & 0.306 & 31.13 & 33.64 & 252.13 \\
    \ding{51} & \ding{51} & \ding{55} & 9.38E-05 & 0.724 & 0.311 & 30.83 & 30.04 & 216.78 \\
    \ding{51} & \ding{51} & \ding{51} & \textbf{8.85E-05} & \textbf{0.734} & \textbf{0.299} & \textbf{29.07} & \textbf{29.74} & \textbf{189.77} \\
    \bottomrule
    \end{tabular}}
\caption{Ablation results of our proposed training components on TikTok dataset. Our proposed~\sysname~utilizing all proposed modules shows the best performance across all image and video metrics. The highest performance is highlighted in bold.}
\label{tab:ablation}
\end{table}
\\
\noindent\textbf{Quantitative Results}
Table~\ref{tab:ablation} quantitatively demonstrates the effectiveness of each proposed method.
The first two rows demonstrate that using the APPA layer to align appearance and pose significantly improves both image and video-level metrics, supporting the notable texture quality enhancement as evidenced in~\cref{fig:APadaptation}.
Additionally, the temporal ControlNet and PTM have enhanced the temporal consistency of the generated videos, contributing to 
quantitative performance improvements, as shown in the last two rows of Table~\ref{tab:ablation}.
\section{Conclusion}
We present~\sysname, a novel human image animation diffusion model with temporally consistent pose guidance. Our approach adapts the latent space between appearance and poses for effective utilization of the pre-trained pose ControlNet and alleviates errors from the off-the-shelf pose extractor by incorporating the temporal module into ControlNet.
Moreover, during inference, our temperature map derived from the input pose sequence promises a consistent background of the output video. 
Experimental results on both challenging TikTok dataset and an extension to the unseen domain using Bizarre pose dataset promise the superiority and generalizability of our method in animating human images. 
Even so, the potential misuse of such videos as deep fakes highlights the ongoing need for research into verification models that can detect traces of generative models.
\appendix
\newcommand{\VidCompTiktok}{\text{`comparison\_tiktok.mp4'}}
\newcommand{\VidCompOnline}{\text{`comparison\_online.mp4'}}



\clearpage
\noindent\textbf{Acknowledgments}
This work was supported by Institute for Information \& communications Technology Promotion(IITP) grant funded by the Korea government(MSIT) (No.RS-2019-II190075 Artificial Intelligence Graduate School Program(KAIST)) and the National Supercomputing Center with supercomputing resources including technical support (KSC-2023-CRE-0445). Finally, we thank all researchers at NAVER WEBTOON Corp.

\bibliographystyle{splncs04}
\bibliography{main}

\clearpage
\appendix
\begin{center}
    \Large
    \textbf{Supplementary Material}
\end{center}

\begin{figure}[ht!]
    \centering
    \vspace{-0.5cm}
    \includegraphics[width=1.0\linewidth]{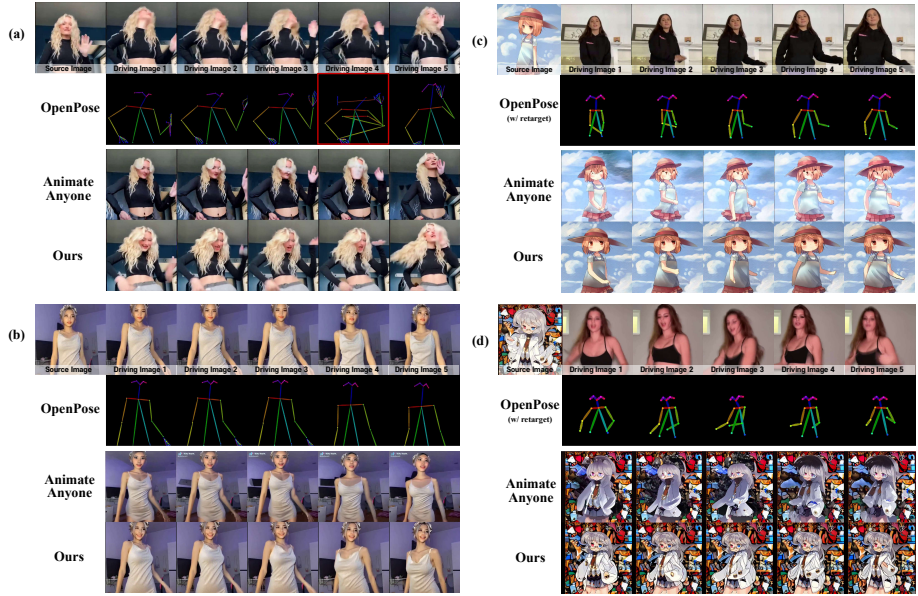}
    \caption{Qualitative comparison of AnimateAnyone~\cite{hu2024animate} and~\sysname~ on TikTok dataset (a and b) and Bizarre dataset (c and d). The erroneous pose is highlighted in red.
    }
    \vspace{-0.5cm}
    \label{fig:supple_qual_moore_merge}
\end{figure}
\section{Comparison with AnimateAnyone}
\begin{table}[ht!]
\vspace{-0.8cm}
\centering
\begin{tabular}{l|cccc|cc} 
\toprule
Method        & L1 $\downarrow$      & SSIM $\uparrow$ & LPIPS $\downarrow$ & FID $\downarrow$  & FID-VID $\downarrow$ & FVD $\downarrow$    \\ 
\hline
AnimateAnyone~\cite{hu2024animate}$^*$ & 1.07E-04 & 0.692 & 0.366 & 28.75 & 42.40   & 341.07  \\
\textbf{\sysname} (\textit{w/ mm1}) & \textbf{8.85E-05} & \textbf{0.734} & \textbf{0.299} & \textbf{29.07} & \textbf{29.74}   & \textbf{189.77}  \\
\bottomrule
\end{tabular}
\vspace{0.2cm}
\caption{Quantitative comparison with AnimateAnyone~\cite{hu2024animate} on TikTok dataset. 
To evaluate AnimateAnyone, we use re-implemented code$^*$.
All methods are trained on TikTok training dataset. We highlight the best performance in bold. }
\vspace{-0.8cm}
\label{tab:supple_quan}
\end{table}
%

In this section, we compare our~\sysname~to the recent human image animation method, AnimateAnyone~\cite{hu2024animate}, which also uses OpenPose as the input pose condition.
Since the AnimateAnyone does not have an official code, we trained re-implemented version of AnimateAnyone
\footnote{\label{Moore}AnimateAnyone(Re-implemented): \href{https://github.com/MooreThreads/Moore-AnimateAnyone}{https://github.com/MooreThreads/Moore-AnimateAnyone}} 
on TikTok training dataset.

Table~\ref{tab:supple_quan} shows the quantitative results of both methods. 
Our model outperforms the re-implemented AnimateAnyone in all performance metrics. 
~\cref{fig:supple_qual_moore_merge} shows the qualitative comparison of TikTok and bizarre dataset. 
As for AnimateAnyone, we observe the distorted face region under inaccurate pose condition (\cref{fig:supple_qual_moore_merge} (a)), and the black object flickering in the background (\cref{fig:supple_qual_moore_merge} (b)).
In contrast, the results of~\sysname~shows well-preserved faces and clean, stable backgrounds. 
~\cref{fig:supple_qual_moore_merge} (c)~and~\cref{fig:supple_qual_moore_merge} (d)~illustrates the results on Bizarre dataset.
\cref{fig:supple_qual_moore_merge}~shows AnimateAnyone fails to preserve identities such as faces or clothing, while our~\sysname~ generalizes well to unseen domain.

\section{Additional Videos for Visual Comparison}
\subsubsection{Qualitative Comparison on TikTok Dataset}
To demonstrate the effectiveness of~\sysname~and to compare our approach with human image animation baselines,
we upload videos on our project page
\footnote{\label{proj}TCAN project page: \href{https://eccv2024tcan.github.io}{https://eccv2024tcan.github.io}} 
, which showcases generated videos where a source image is the first frame of each driving video. 
The selected frame from the video is attached in~\cref{fig:supp_qual_video} (a). 
It is evident from the video that~\sysname~yields the most realistic results compared to the others.
Among all baselines, MagicAnimate stands out as the closest competitor. 
MagicAnimate's advantage comes from its use of DensePose for guidance, providing detailed shape information from the driving video that aligns with the shape of the source image's foreground object.
In contrast, our~\sysname~achieves remarkable performance by utilizing sparse pose information only.

\subsubsection{Qualitative Comparison on Online-Collected Images}
\label{sec:vid_online}
\begin{figure}[t!]
    \centering
    \includegraphics[width=1\linewidth]{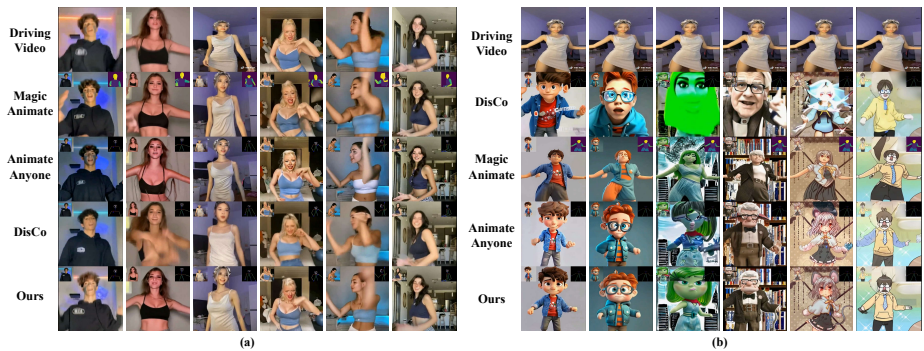}
    \caption{Additional~\sysname~results on (a) TikTok dataset and (b) various animation characters. We select frames from each video at the same time and compare them. The full video can be viewed on our project page.
    }
    \label{fig:supp_qual_video}
\end{figure}
\begin{figure}[t!]
    \centering
    \includegraphics[width=1\linewidth]{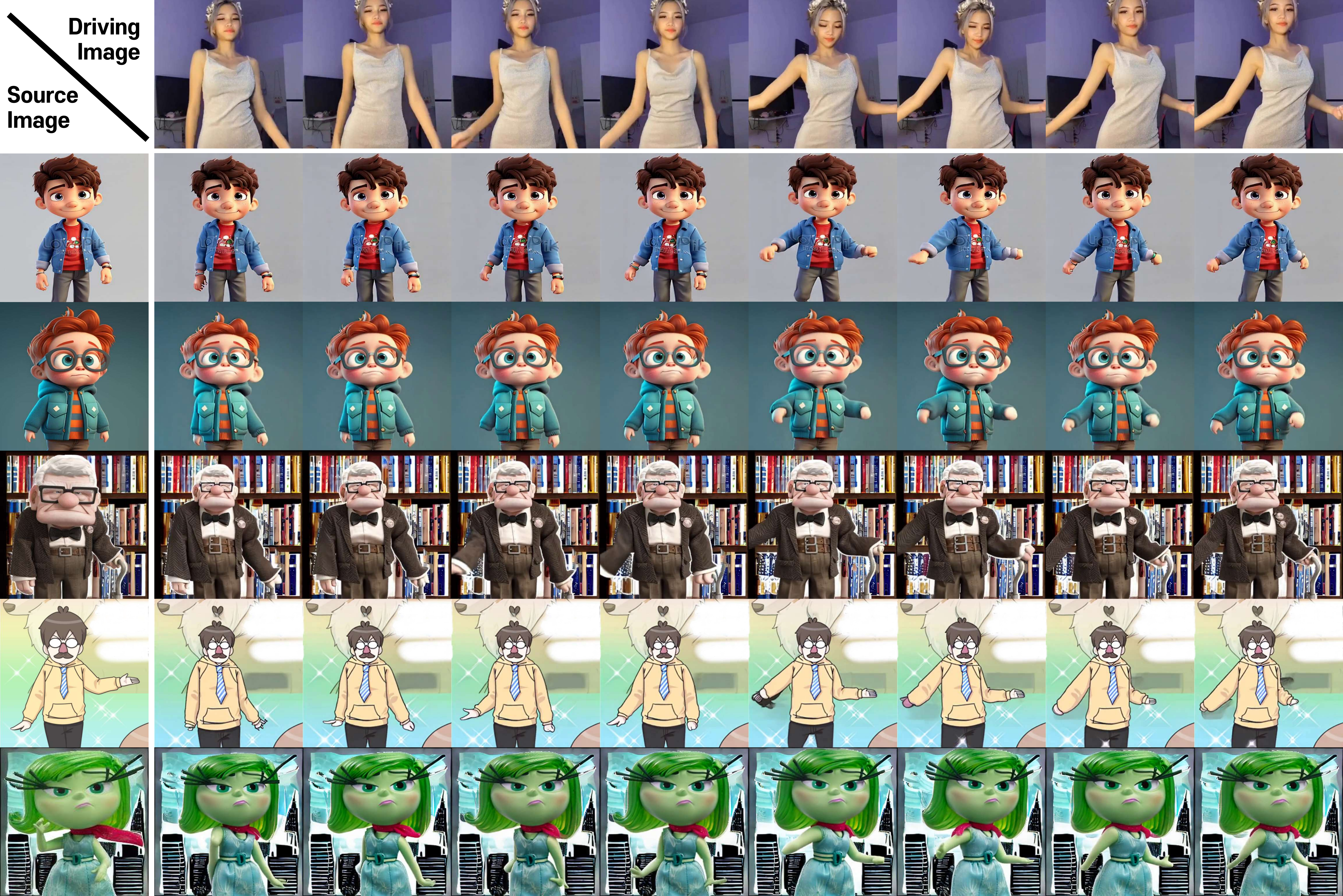}
    \caption{Additional~\sysname~results on various animation characters. 
    }
    \label{fig:supp_qual_retarget}
\end{figure}
In this section, we show additional results on the online-collected animation characters.
We have collected characters with diverse body proportions and styles, and use them as source images. 
As depicted in~\cref{fig:supp_qual_retarget}, 
our methodology follows the motion of the foreground object in driving videos while effectively maintaining the identity and background of the source image. 
Furthermore, we upload generated videos using animation characters on our project page
for qualitative comparisons with baseline methods (\textit{i.e.,} DisCo, MagicAnimate, and re-implemented AnimateAnyone). 
The sampled frames from the video are shown in~\cref{fig:supp_qual_video} (b).  
As shown in the the figure, 
DisCo exhibits significant distortion of the source image's identity, and MagicAnimate changes the character's body proportions, leading to 
an identity shift. 
In contrast, our~\sysname~effectively follows the motion of driving videos with diverse body proportions and styles, even when trained solely on the TikTok dataset.



\section{Detailed Description of Pose Retargeting}
\definecolor{functioncolor}{RGB}{236, 82, 75}
\definecolor{commentcolor}{RGB}{110,154,155}   
\newcommand{\PyFunction}[1]{\ttfamily\textcolor{functioncolor}{#1}}  
\newcommand{\PyComment}[1]{\ttfamily\textcolor{commentcolor}{\# #1}}  
\newcommand{\PyCode}[1]{\ttfamily\textcolor{black}{#1}} 
\begin{algorithm}[ht!]
    \SetAlgoLined
    \scriptsize{
        \PyFunction{def} pose\_retarget(cur\_kp, src\_kp, init\_kp, kp\_mapper): \\
        \Indp   
        \PyComment{cur\_kp : 18 keypoint coordinates from the current frame.} \\
        \PyComment{src\_kp : 18 keypoint coordinates from the source image.} \\
        \PyComment{init\_kp : 18 keypoint coordinates from the initial frame.} \\
        \PyComment{kp\_mapper : A mapper storing the relationship between two connected keypoints.} \\ 
        \quad \\
        \PyComment{Calculate the ratio of the lengths between the keypoints of the initial and the current frame.} \\
        curinit\_ratio = [1] * 18 \\
        for kp1 in range(2, 18): \\
            \Indp
            kp2 = kp\_mapper[kp1] \\
            cur\_length = get\_length(cur\_kp[kp1], cur\_kp[kp2]) \\
            init\_length = get\_length(init\_kp[kp1], init\_kp[kp2]) \\
            curinit\_ratio[kp1] = cur\_length / init\_length \\
            \Indm
        \quad \\
        \PyComment{First adjusting the length between the neck(1)-nose(0).} \\
        neck\_kp = 1 \\
        nose\_kp = kp\_mapper[neck\_kp] \\
        src\_length = get\_length(src\_kp[neck\_kp], src\_kp[nose\_kp]) \\
        cur\_length = get\_length(cur\_kp[neck\_kp], cur\_kp[nose\_kp]) \\
        ratio = cur\_length / src\_length \\
        diff\_vector = get\_diff\_with\_ratio(cur\_kp[neck\_kp], cur\_kp[nose\_kp], ratio) \\
        cur\_kp[nose\_kp] += diff\_vector \\
        \quad \\
        \PyComment{Move all subkeypoints of the nose keypoint (i.e., eyes and ears)} \\
        cur\_kp = move\_subkeypoints(cur\_kp, nose\_kp, diff\_vector) \\
        \quad \\

        \PyComment{Traverse and move the remaining keypoints and subkeypoints based on the length of the nose-neck.}\\
        for kp1 in range(2, 18): \\
            \Indp
            kp2 = kp\_mapper[kp1] \\
            src\_length = get\_length(src\_kp[kp1], src\_kp[kp2]) \\
            cur\_length = get\_length(cur\_kp[kp1], cur\_kp[kp2]) \\
            ratio = src\_length / cur\_length * curinit\_ratio[kp2] \\
            diff\_vector = get\_diff\_with\_ratio(cur\_kp[kp2], cur\_kp[kp2], ratio) \\
            cur\_kp[kp2] += diff\_vector \\
            cur\_kp = move\_subkeypoints(cur\_kp, kp2, diff\_vector) \\
            \Indm
        retargeted\_cur\_kp = cur\_kp \\
        \PyFunction{return} retargeted\_cur\_kp
        
    }
    \caption{Pseudo-Code for Pose-Retargeting}
\label{algo:pose_retargeting}
\end{algorithm}
As discussed in section~\ref{sec:bizarre}, we apply the pose re-targeting algorithm to align the ratio of foreground objects of driving images to that of the source image. Specifically, we fix the `neck' keypoint of the driving image and adjusted the position of the `nose' so that the distance between the `nose' and `neck' keypoints in the source image matched the corresponding distance in the driving image. The keypoints connected to the `nose' were moved by the same amount. Similarly, we traverse through the remaining keypoints, adjusting their positions to match the distances between the keypoints in the source image. We provide a detailed description of Pose-Retargeting in Algorithm~\ref{algo:pose_retargeting} as pseudo-code.

\end{document}